\documentclass[lettersize,journal]{IEEEtran}

\usepackage{mathptmx} % assumes new font selection scheme installed
\usepackage{times} % assumes new font selection scheme installed
\usepackage[dvipdfmx]{graphicx}
\usepackage{url}
\usepackage{graphicx}
\usepackage[figurename=Fig.\ ]{caption}
\usepackage{epsfig} % for postscript graphics files
\usepackage{cite}
\usepackage{color}
\usepackage{booktabs}  % 
\usepackage{amsmath}
\usepackage{amssymb}
\usepackage{indentfirst}
\usepackage{algorithm}
\usepackage{algorithmicx}
\usepackage{algpseudocode}
\usepackage{caption}
\usepackage{subcaption}
\usepackage{comment}
\usepackage{color,soul}
\usepackage{ifthen}
\usepackage[table,xcdraw]{xcolor}
\usepackage{hyperref}
\usepackage{animate}
\graphicspath{ {./img/} }

\title{\LARGE \bf
Text-driven object affordance for guiding grasp-type recognition in multimodal robot teaching
}

\author{
Naoki Wake$^{1}$,
Daichi Saito$^{2}$,
Kazuhiro Sasabuchi$^{1}$,
Hideki Koike$^{1}$,
and Katsushi Ikeuchi$^{1}$
\thanks{$^{1}$Applied Robotics Research, Microsoft, Redmond, WA 98052, USA
        {\tt\small naoki.wake@microsoft.com}
        $^{2}$Department of Computer Science, Tokyo Institute of Technology, Meguro, Tokyo 1528550, Japan
        }%
}
\begin{document}
\maketitle
\thispagestyle{empty}
\pagestyle{empty}

\begin{abstract}
This study investigates how text-driven object affordance, which provides prior knowledge about grasp types for each object, affects image-based grasp-type recognition in robot teaching. The researchers created labeled datasets of first-person hand images to examine the impact of object affordance on recognition performance. They evaluated scenarios with real and illusory objects, considering mixed reality teaching conditions where visual object information may be limited. The results demonstrate that object affordance improves image-based recognition by filtering out unlikely grasp types and emphasizing likely ones. The effectiveness of object affordance was more pronounced when there was a stronger bias towards specific grasp types for each object. These findings highlight the significance of object affordance in multimodal robot teaching, regardless of whether real objects are present in the images. Sample code is available on \href{https://github.com/microsoft/arr-grasp-type-recognition}{GitHub}.
\end{abstract}

\section{Introduction}
Robot grasping has been a major issue in robot teaching for decades \cite{cutkosky1990human,kang1997toward}.
Because robot grasping determines the positional relationship between a robot's hand and an object, grasping objects suitable for the given environment is critical for efficient and successful manipulations after grasping. Recent studies have focused on learning-based end-to-end robot grasping \cite{lenz2015deep,morrison2018closing,jiang2011efficient,bousmalis2018using,redmon2015real,yu2018robotic}, where contact points or motor commands are estimated from visual input. However, a desired grasp differs depending on the type of manipulation to be achieved, even for the same target object. 
While such grasp uncertainty can be addressed in an automatic manner using an advanced robot control method (e.g., ~\cite{tutsoy2021novel}), a simpler approach can be employed in the context of robot teaching, where a human teaches the robot how to grasp through a demonstration.

\begin{figure}[tb]
	\centering
	\includegraphics[width=\linewidth]{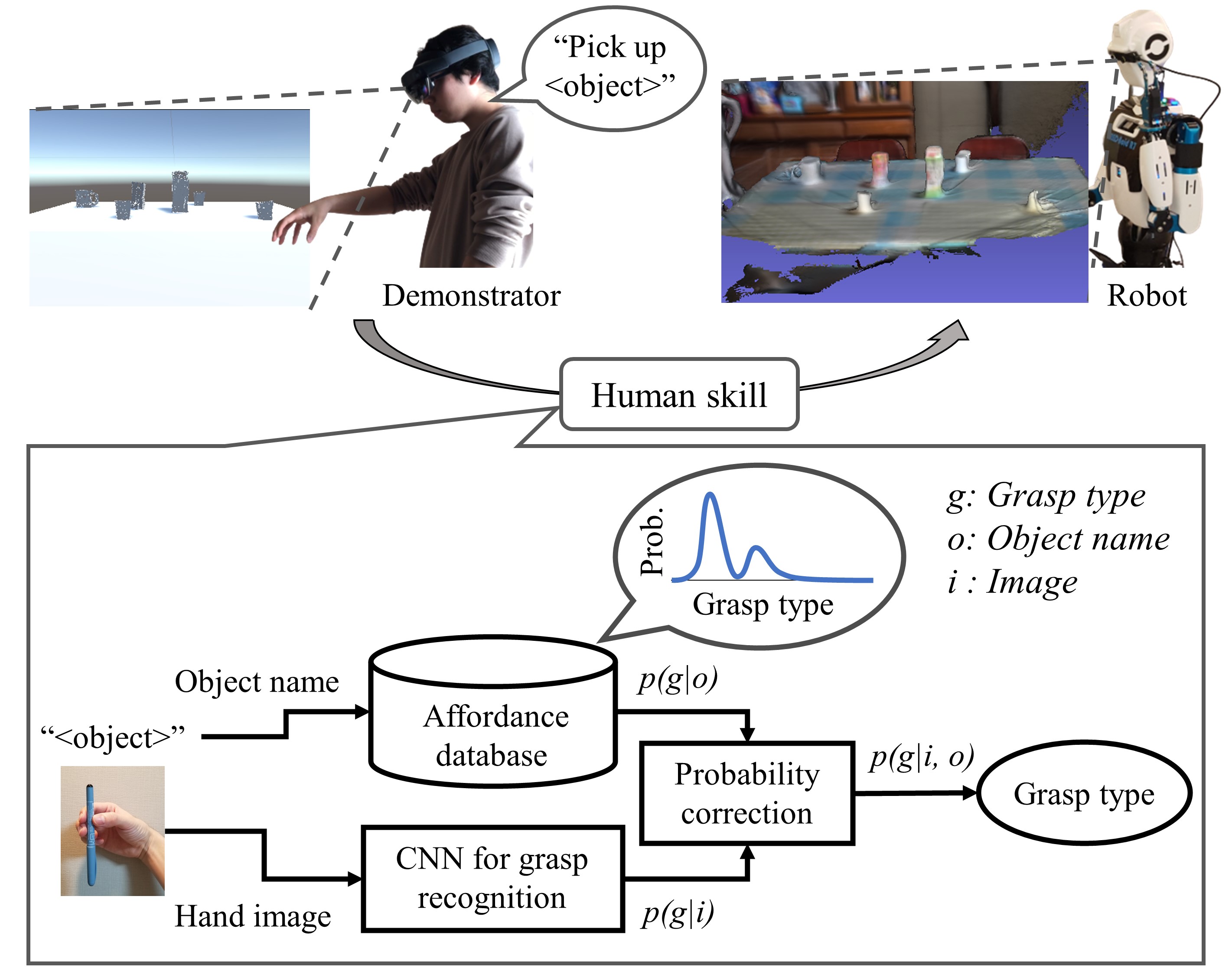}
	\caption{Conceptual diagram of robot teaching. (Top) Head-mounted device provides first-person images during a demonstration with verbal instructions (modified version of image from \cite{saito2021contact}). The demonstrations are transferred to a robot in the form of a skill set, which includes a grasp type. (Bottom) Proposed pipeline for grasp-type recognition leveraging object affordance. The pipeline estimates the grasp type from the pairing of an object name and an image of a hand grasping that object. Object affordance is searched from an affordance database using text matching (modified image in \cite{wake2020grasp}).}
	\label{fig:fig9A}
\end{figure}

We have been developing a platform to teach a robot ``how to grasp and manipulate an object'' through multimodal human demonstrations~\cite{saito2021contact,wake2021learning, wake2020verbal,sasabuchi2020task,saito2022task} (Fig.~\ref{fig:fig9A}). The demonstration is accompanied by verbal instructions and captured by a head-mounted device (HMD). The user demonstrates object manipulation using either physical objects or illusory objects superimposed by the HMD on the demonstrator's hands. According to the definition of Milgram et al.~\cite{milgram1994taxonomy}, we refer to the later setup as mixed reality (MR). 
Assuming such a multimodal teaching system, this paper focuses on recognizing grasp types based on the name of the object and the first-person image at the time of grasping.

The problem of recognizing grasp types from human grasping images is not new. However, because grasp-type recognition has developed in the context of computer vision, most existing research is image-based (e.g.,~\cite{cai2018understanding, corona2020ganhand, rogez2015understanding}). When considered in terms of multimodal robot teaching, further questions arise. 1) How can linguistic input be utilized? 2) In what situations is linguistic input more advantageous? 3) Is linguistic input useful even in a challenging situation, such as MR, where images are not projected? Although these are practical and important questions in robot teaching, to the best of our knowledge, no previous studies have addressed these issues.

In many cases, an object name is known to be associated with the possible grasp types \cite{helbig2010action,feix2014analysis,cini2019choice,corona2020ganhand}. Based on this association, we have previously proposed a pipeline that leverages a prior distribution of grasp types to improve a convolutional neural network (CNN)-based image recognition~\cite{wake2020grasp}(Fig.~\ref{fig:fig9A} bottom). We refer to the prior distribution as object affordance, a concept proposed by Gibson \cite{gibson1966senses}. In the pipeline, appropriate object affordance was searched from an affordance database using text matching.
Although our preliminary experiments suggested that the object affordance is a promising solution to leverage a user's linguistic input for grasp-type recognition, its effectiveness has not been fully understood due to the lack of a dataset.

This study aimed to investigate the role of object affordance for multimodal grasp-type recognition. To this end, we prepared a large first-person grasping image dataset containing a wider range of labeled grasp types and household objects. We tested the pipeline with two types of affordances, which reflect one or both of the likeliness and unlikeliness of each grasp type. The experiments showed that object affordance guides CNN recognition in two ways: 1) excluding unlikely grasp types from the candidates and 2) enhancing likely grasp types among the candidates. 
Additionally, the ``enhancing effect'' was more pronounced with a greater grasp-type bias for each object in a test dataset. 
Furthermore, we tested the pipeline for recognizing mimed grasping images (i.e., images of a hand grasping an illusory object), assuming that a real object may be absent in some situations (e.g., teaching in MR). Similar to the experiment with real grasping images, object affordance proved to be effective for mimed grasping images. Additionally, the CNN recognition for the mimed images exhibited lower performance compared to its recognition for real grasping, indicating the importance of the presence of real objects in image-based recognition.

The contributions of this study are 1) demonstrating the effectiveness of the object affordance in guiding grasp-type recognition both with and without the real objects in images, 2) demonstrating the conditions under which the merits of object affordance are pronounced, and 3) providing a dataset of first-person grasping images labeled with the possible grasp types for each object.

The remainder of this paper is organized as follows. Section \ref{related_work} provides an overview of the proposed pipeline and related works. Section \ref{evaluation} describes the experiments conducted with and without real objects. Finally, Section \ref{conclusion} summarizes the results of the study and describes future work.

\section{System overview} \label{related_work}
\subsection{Grasp taxonomy}
There are two main approaches to analyzing human grasping from a single image: 1) using hand poses of grasping \cite{kokic2020learning,kokic2019learning,hasson2019learning} and 2) using a specific grasp taxonomy \cite{feix2014analysis,rogez2015understanding,cai2018understanding,cai2015scalable,huang2015we,saudabayev2018human,feix2014analysis2, arapi2021understanding}. 
% arapi2021understanding
Each approach has its own advantages. Hand pose analysis in 3D space enables measurement of object states, such as posture \cite{kokic2019learning} and grasping area \cite{kokic2020learning}. Meanwhile, taxonomy analysis enables human grasps to be represented as discrete intermediate states that focus on the pattern of the fingers in contact. 
This study aimed to recognize grasp types from human behavior as an extension of taxonomy-based studies. We employed the taxonomy by Feix et al., which contains 33 grasp types \cite{feix2015grasp}.

\subsection{Dataset of human grasps}
Building a realistic dataset of human hand shapes while manipulating objects will contribute to the study of human grasping. Some studies collected joint positions using wired sensors \cite{garcia2018first}, a data glove \cite{lin2014grasp}, and model fittings \cite{hasson2019learning,hampali2020honnotate}. Another study created a dataset of hand--object contact maps obtained using thermography \cite{brahmbhatt2019contactdb}. Additionally, taxonomy-based studies have created datasets annotated with grasp types \cite{rogez2015understanding,cai2018understanding,saudabayev2018human,bullock2015yale}. For example, Bullock et al. collected a dataset containing first-person images of four workers \cite{bullock2015yale}. 

Despite the variety of datasets available for grasp-type recognition, they could not be directly applied to our study because they do not aim to cover the possible grasp types associated with an object. Although there exists a pseudo-image dataset focusing on object-specific affordance \cite{corona2020ganhand}, there is no dataset that provides the actual grasping images. In contrast, the uniqueness of the dataset in this study is that it aimed to cover the possible grasp types for each object while providing RGB images of real human grasps. Additionally, the objects were selected from common household objects (see Section \ref{data_preparation} for details).

\subsection{Object affordance}\label{object_affordance}
The originality of this research is that we introduce object affordance obtained by searching a database by an object name. Although several studies have reported the effectiveness of using multi-modal cues for grasp-type recognition \cite{cai2018understanding,yang2015robot}, the understanding of the effectiveness of linguistically-driven object affordance is still limited in the context of multimodal robot teaching. 

In concrete implementation, object affordance was represented by a dictionary with object names as keys. When an object name was input to the pipeline, the object affordance corresponding to the object name was retrieved by searching the dictionary (Fig.~\ref{fig:fig9A}). Note that this affordance database was not acquired automatically but was assumed to be added and modified by the user according to each application.

\subsubsection{Definition of object affordance}
Prediction of affordance has become an active research topic in the cross-domain of robotics and computer vision. Affordance, which is generally regarded as an opportunity for interaction in a scene, has been defined in different ways depending on the problem to be solved. For example, in the computer vision research using deep learning, affordances have been formulated as a type of label in semantic segmentation tasks \cite{brahmbhatt2019contactdb,do2018affordancenet,porzi2016learning,lau2016tactile,roy2016multi}. 
In robotics research, affordance is a topic of the task-dependent object grasping problem, which is referred to as task-oriented grasping (TOG)~\cite{kokic2020learning}. 
In the context of TOG, affordance is defined as the possible tasks (e.g., cut and poke) allowed for an object \cite{kokic2017affordance,fang2020learning,song2015task,bohg2013data}. 

In this study, object affordance was defined for each object as ``a distribution of the possible grasp types associated with the object's name.'' This definition is similar to TOG in that it considers affordance to be object-specific. However, our definition focuses on the grasp types and does not scope the information on the possible tasks following the grasps.

\subsubsection{Types of object affordance evaluated}
\begin{figure}
    \centering
    \includegraphics[width=\linewidth]{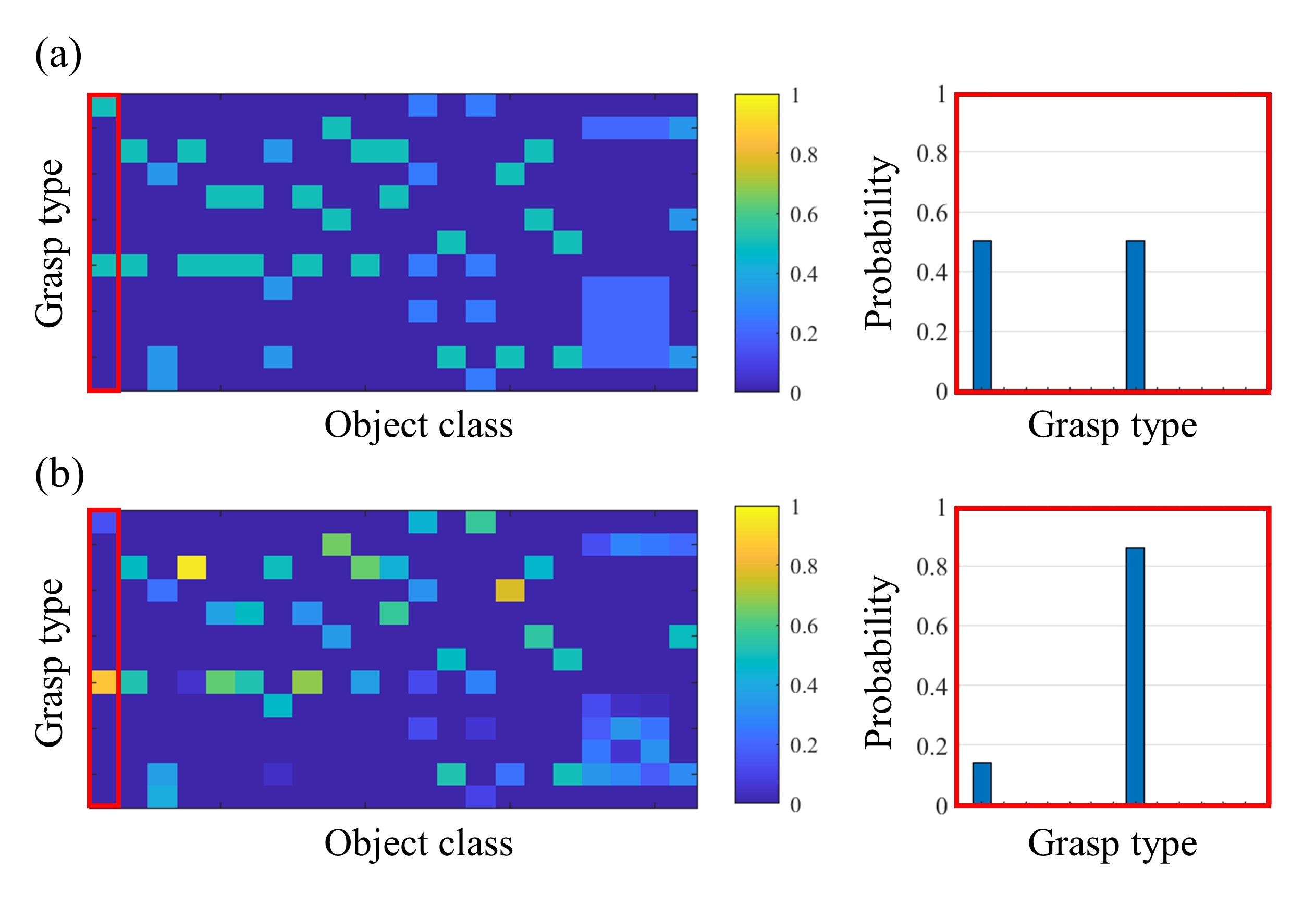}
    \caption{Examples of object affordance calculated from a sub-dataset: (a) example of uniform affordance and (b) example of varied affordance. Refer to Fig. \ref{fig:fig2} for the order of grasp types and object classes.}
    \label{fig:fig3}
\end{figure}

The experiments in Section~\ref{evaluation} evaluate the role of object affordance using sub-datasets that were sampled from the created dataset, which was labeled with the possible grasp types for each object (see Section~\ref{data_preparation} for details). While testing the proposed pipeline (Fig.~\ref{fig:fig9A}), an affordance database was created for each sub-dataset based on the grasp-type labels found in the sub-dataset. We prepared two types of affordances for each object (Fig.~\ref{fig:fig3}):

\begin{itemize}
    \item \textit{Varied affordance} was calculated as a normalized histogram of the labeled grasp types for each object. 
    
    \item \textit{Uniform affordance} was calculated by flattening the non-zero values in the histogram. 

\end{itemize}

While the varied affordance contains information regarding the likeliness and unlikeliness of grasping, the uniform affordance only contains information regarding the unlikeliness of grasping.

\subsection{Convolutional neural network with object affordance} \label{cnn_affordance}
We formulated grasp detection by fusing a CNN with object affordance (Fig.~\ref{fig:fig9A}) as follows. The image, object name, and grasp type are denoted as $i$, $o$, and $g$, respectively. Assuming the output of the CNN to be a probability of each grasp type $g$ given an image $i$, we represent the output of the CNN as $p(g \mid i)$. Additionally, based on the definition of object affordance (i.e., a distribution of the possible grasp types associated with the object's name), we represent the object affordance as $p(g \mid o)$. Herein, we focused on deriving the probability of each grasp type given both an image and an object name (i.e., $p(g \mid i,o)$) from these conditional probabilities, $p(g \mid i)$ and $p(g \mid o)$. Assuming that $p(i)$ and $p(o)$ are independent, the following equation holds based on mathematical formulas:

\begin{equation}\label{eq:affodance}
\begin{split}
p(g\mid i,o)&=\frac{p(i,o \mid g)\,p(g)}{p(i,o)} \\
            &=\frac{p(i \mid g)\,p(o \mid g)p(g)}{p(i)\,p(o)} \\
            &=\frac{p(g \mid i)\,p(g \mid o)}{p(g)}
\end{split}
\end{equation}
Hence, the conditional probability distribution $p(g \mid i,o)$ can be estimated from the available distributions $p(g \mid i)$, $p(g \mid o)$, and $p(g)$. Finally, the grasp type can be determined as that which maximizes $p(g \mid i,o)$.
A reasonable interpretation of this equation is that the grasp-type recognition based on object name and image can be approximated by a measure that considers the predictions based on the object name and image respectively, and the rarity of the grasp type (i.e., $1/p(g)$).

A CNN network was obtained by fine-tuning ResNet-101 \cite{he2016deep}. To avoid overfitting, we applied random reflection and translation to images, and randomly shifted the image color in the hue, saturation, value space after every training epoch. The learning was conducted using the Adam optimizer % \cite{kingma2014adam}
and continued until the validation accuracy ceased increasing. %The number of training images was changed for each experiment.

\section{Experiments} \label{evaluation}
\subsection{Scenario 1: with real objects} \label{scenario1}
In this scenario, the demonstration of grasping a real object was provided as a first-person image using an HMD. We assumed that the system could retrieve object affordance from the affordance database using the name of the object mentioned through verbal instructions (e.g., ``Pick up the \textbf{apple}.''). 

\begin{figure*}[tb]
	\centering
	\includegraphics[width=\linewidth]{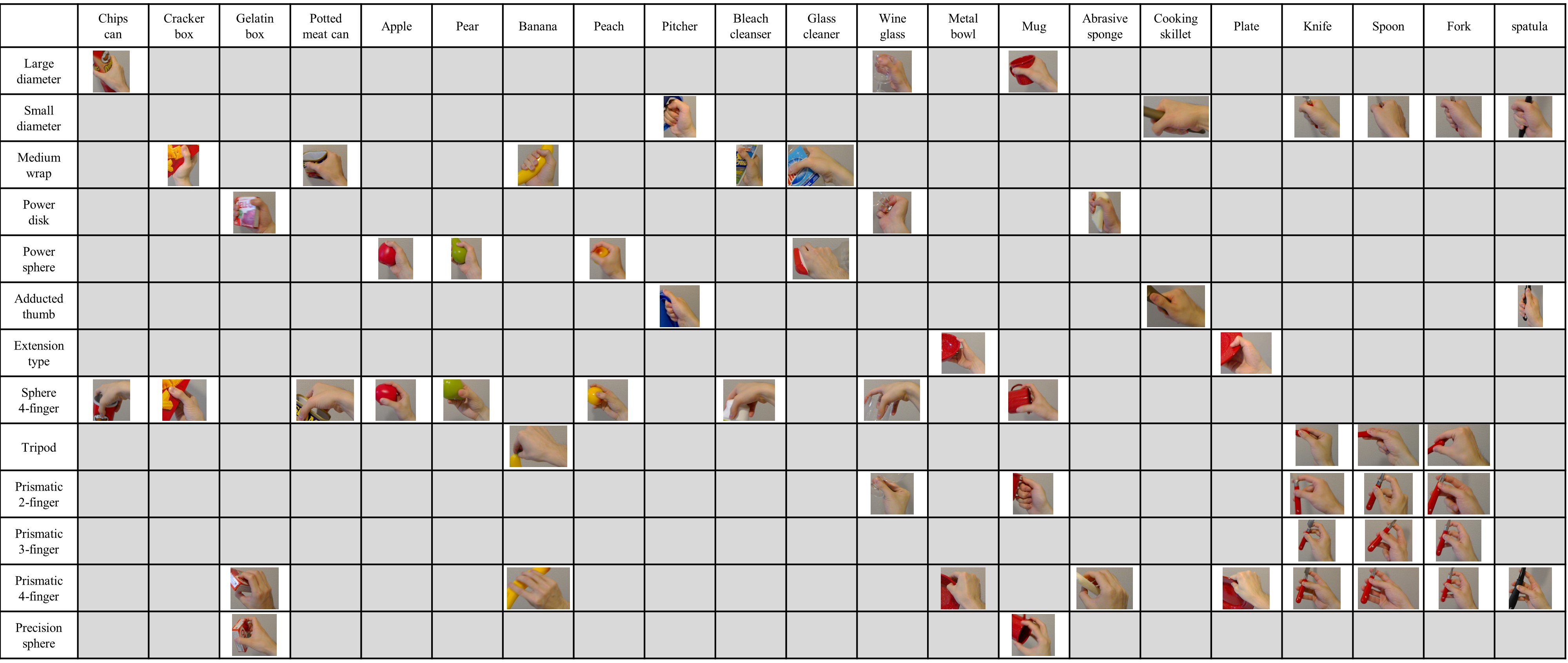}
	\caption{Grasp types assigned to Yale-CMU-Berkeley (YCB) objects. Images were selected from the database to demonstrate examples of grasping.}
	\label{fig:fig2}
\end{figure*}

\subsubsection{Data preparation} \label{data_preparation}
Demonstrations are frequently recorded by an HMD in MR-based robot teaching (e.g., ~\cite{guan2019novel, saito2021contact}). Even for robot teaching in the physical world, first-person images provided by the demonstrator are preferred over third-person images owing to their ability to avoid self-occlusion. Therefore, we required a dataset of first-person images labeled with the possible grasp types and object names. Because we were not able to find any existing dataset that met these requirements, we created one. 

The images were captured by a HoloLens2 sensor \cite{hololens}. We used this sensor because it is a commercially available sensor that can capture first-person images without the need for hand-made attachments. The target object was chosen from the Yale-CMU-Berkeley (YCB) object set \cite{calli2015ycb}, which covers common household items. We employed this object set because it has been used as a benchmark for many robotic studies. We selected eight items from the food category and 13 items from the kitchen category: chip can, cracker box, gelatin box, potted meat can, apple, banana, peach, and pear; and pitcher, bleach cleanser, glass cleaner, wine glass, metal bowl, mug, abrasive sponge, cooking skillet, plate, fork, spoon, knife, and spatula, respectively. We selected these items to encompass a variety of sizes. 
We prepared two datasets to avoid the overestimation of the performance of the network due to CNN overfitting:

\begin{itemize}
    \item \textit{YCB dataset}: Training dataset containing exactly the same items as the YCB object set.
    
    \item \textit{YCB-mimic dataset}: Testing dataset containing objects that are the same as those in the YCB dataset but differ in terms of color, texture, or shape (e.g., a cracker box from another manufacturer).
\end{itemize}

The datasets were prepared through the following pipeline. Before collecting the images, we manually assigned a set of plausible grasps according to the taxonomy in \cite{feix2015grasp} (Fig. \ref{fig:fig2}). Based on a previous study \cite{lin2015robot}, we focused on 13 grasp types that we believed were achievable for common robot hands. For each object and grasp type, we captured images of a human grasping the object using their right hand. We captured more than 1500 grasping images by varying the arm orientation and rotation as much as possible. Furthermore, to crop the hand regions from the captured images, we applied a third-party hand detector \cite{jsk_ssd} in offline. After manually filtering out the detection errors, 1000 images were randomly collected for each object and grasp type. The following experiments were conducted with sub-datasets that were sampled from the YCB or YCB-mimic dataset.

\subsubsection{Evaluation of dataset size} \label{dataset_size}
Because small datasets lead to underestimation in CNN recognition, we validated the performances of the CNNs trained with different sized sub-datasets of the YCB dataset. We prepared five sub-datasets containing 10, 50, 100, 500, and 1000 images per grasp type. The images were randomly sampled such that a sub-dataset included all the images from the other smaller sub-datasets. The CNNs were tested with sub-datasets of the YCB-mimic dataset. We refer to these sub-datasets as the test datasets. The test datasets were created by randomly sampling 100 images per grasp type. The performances of the CNNs were validated ten times using different test datasets.

\begin{figure}[tb]
	\centering
    \includegraphics[scale=0.32]{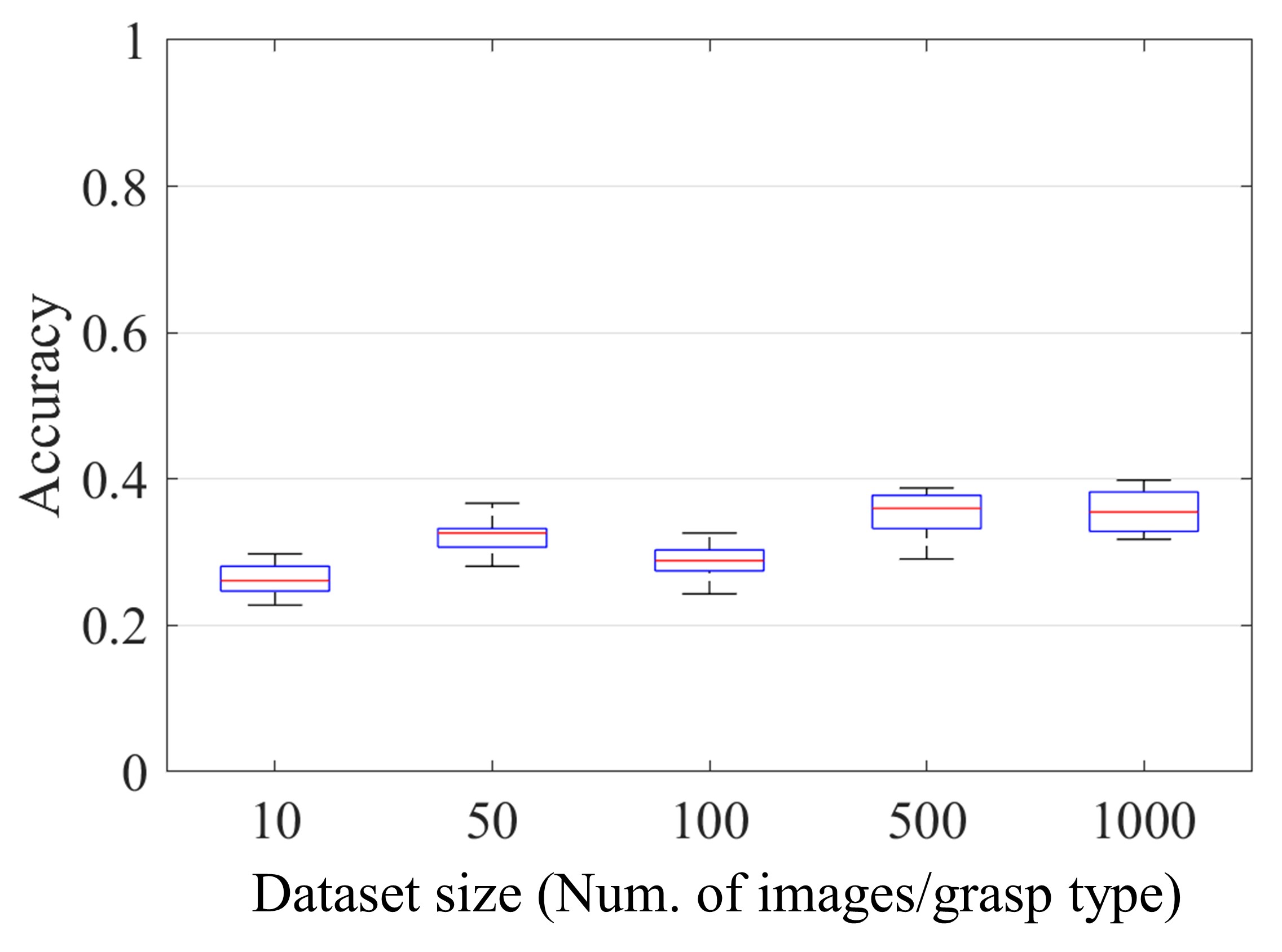}
	\caption{Performances of CNNs trained with different dataset sizes.}
	\label{fig:fig4}
\end{figure}

Fig.~\ref{fig:fig4} shows the result. The CNN performance tended to increase with the size of the dataset and converged above 500 images per grasp type. This result indicates that the YCB dataset is sufficiently large to avoid underestimation due to insufficient images.

\subsubsection{Effect of affordance on recognition}
We evaluated the effectiveness of the proposed pipeline by comparing five methods: the proposed pipeline using 1) varied affordance (i.e., $p(g \mid i,o)$), 2) uniform affordance, 3) only varied affordance (i.e., $p(g \mid o)$), 4) only uniform affordance, and 5) only the CNN (i.e., $p(g \mid i)$). For fair comparison, the same CNN was used for each method. The grasp type that maximizes the probability distribution was chosen. In the case of using only the uniform affordance, the grasp type was randomly selected from the possible grasp types.

The CNN was trained with a sub-dataset of the YCB dataset. Based on the evaluation of the dataset size in Section~\ref{dataset_size}, the sub-dataset was prepared by randomly sampling 1000 images per grasp type.
We compared the performances of five methods applied to a set of 100 test datasets. Each test dataset was created by randomly sampling 100 images per object from the YCB-mimic dataset. 

\begin{figure}[tb]
	\centering
    \includegraphics[scale=0.32]{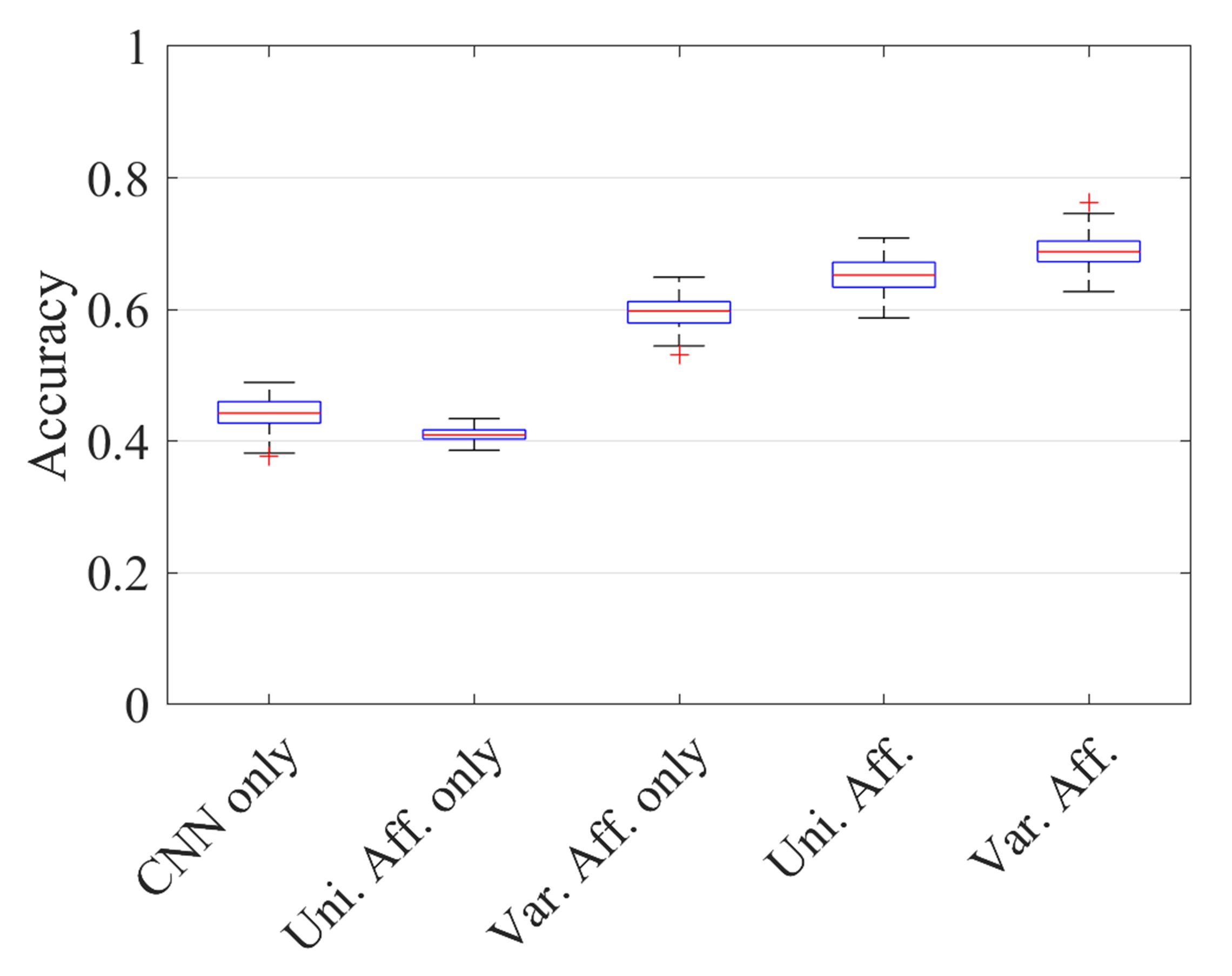}
	\caption{Performance of grasp-type recognition with different pipelines: CNN only (only the CNN), Uni. Aff. only (only uniform affordance), Var. Aff. only (only varied affordance), Uni. Aff. (proposed pipeline using uniform affordance), and Var. Aff. (proposed pipeline using varied affordance).}
	\label{fig:fig5}
\end{figure}

Fig.~\ref{fig:fig5} shows the result. The pipelines combining the CNN and affordance performed better than the CNN-only and affordance-only pipelines. While the proposed affordance exhibited the best performance, the proposed pipeline using uniform affordance was comparable. These increased performances indicate the effectiveness of using affordance for guiding grasp-type recognition.

\begin{figure}[tb]
	\centering
    \includegraphics[width=\linewidth]{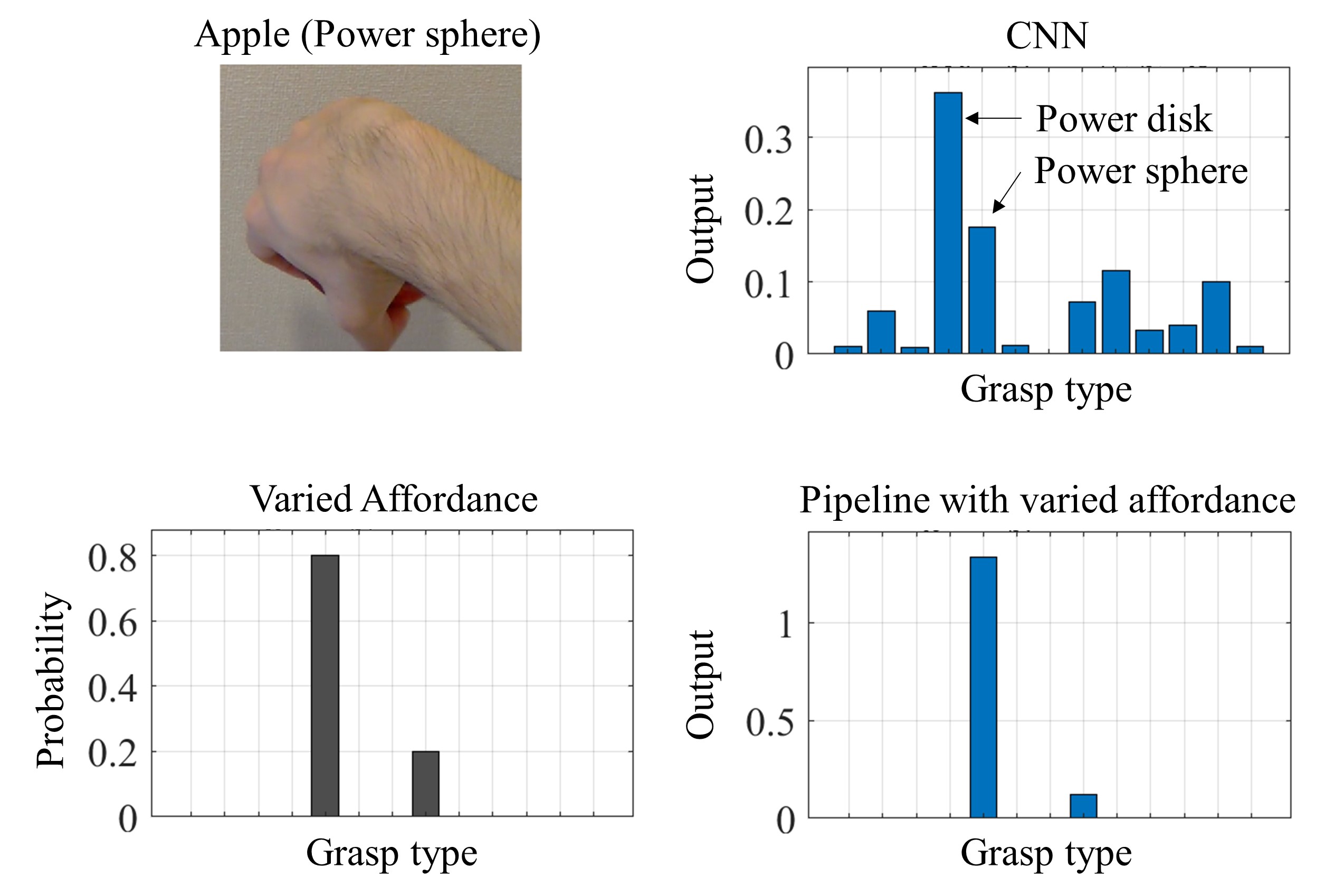}
	\caption{Example of where the CNN failed. The order of grasp types is the same as in Fig.~\ref{fig:fig2}.}
	\label{fig:fig6}
\end{figure}

To elucidate the role of affordance, we examined the cases where the CNN failed whereas the use of varied affordance succeeded (Fig.~\ref{fig:fig6}). In such cases, the CNN did not output the correct grasp as the best candidate, possibly due to finger occlusion; however, it had a small affordance value, resulting in a small likelihood to be the output of the proposed method. As a result, the correct grasping was chosen as the final output. Therefore, evidently, object affordance contributed to excluding the unlikely grasp types from the candidates of the CNN.

\begin{figure}[tb]
	\centering
    \includegraphics[width=\linewidth]{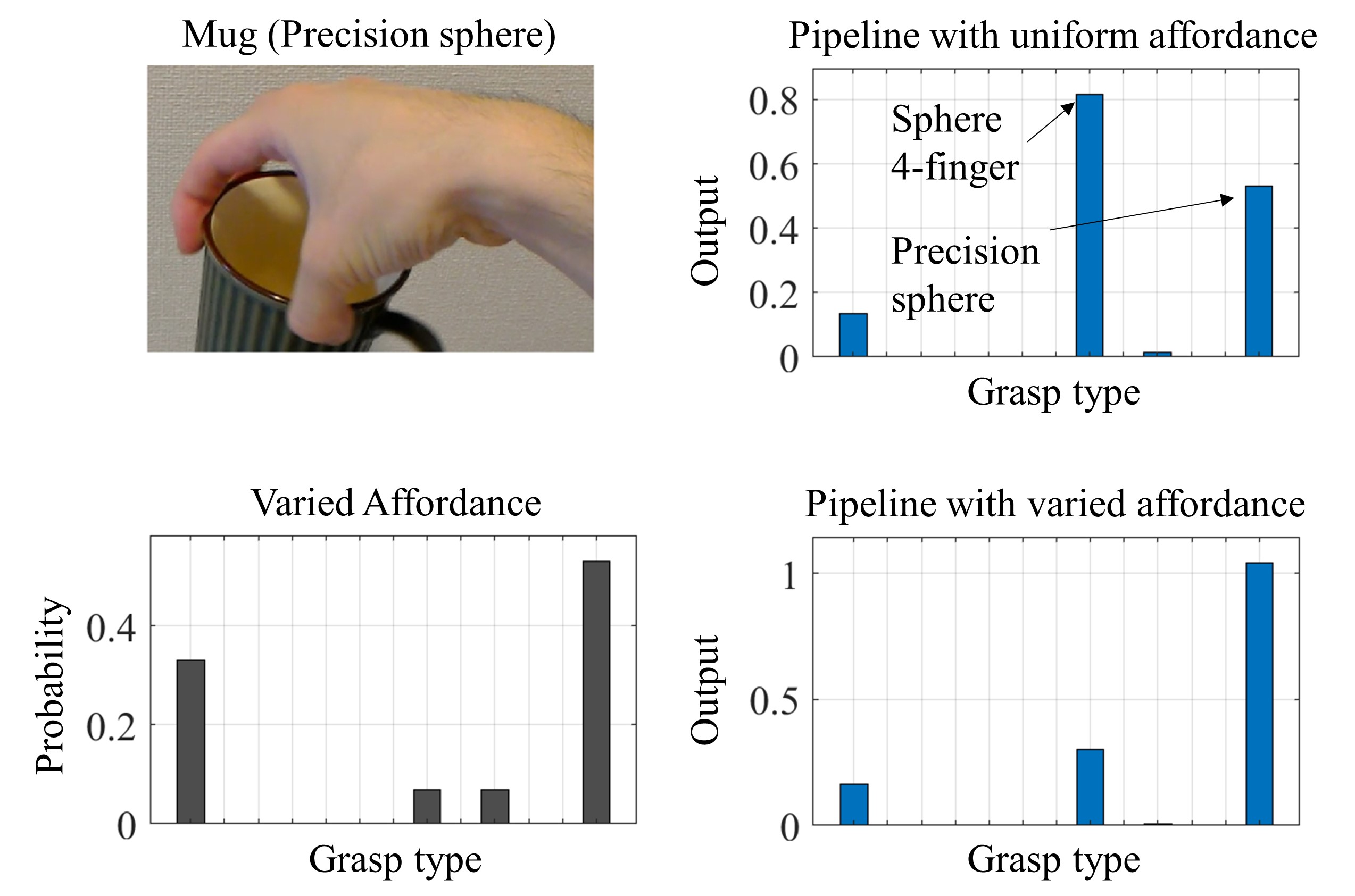}
	\caption{Example of where the proposed pipeline using uniform affordance failed. The order of grasp types is the same as in Fig.~\ref{fig:fig2}.}
	\label{fig:fig7}
\end{figure}

To investigate the advantage of varied affordance over uniform affordance, we examined cases where the use of uniform affordance failed whereas the use of varied affordance succeeded (Fig.~\ref{fig:fig7}). In these cases, the pipeline outputted the correct grasping by employing varied affordance. Therefore, evidently, varied affordance contributed to enhancing the grasp types that were likely for an object.


\subsubsection{Enhancing effect of varied affordance}
After observing the enhancing effect of the object affordance, based on information theory, we hypothesized that the effect would be stronger with 
%higher grasp-type heterogeneity. 
greater grasp-type bias for each object in a test dataset (i.e., grasp-type heterogeneity).
To test this hypothesis, we evaluated the effect of grasp-type heterogeneity on recognition.

We used the same 100 test datasets that were prepared for the comparison experiment. The degree of grasp-type heterogeneity, $h$, was defined for each test dataset by the following equation:
\begin{equation}
h=\frac{1}{N}\sum_{i=1}^{N}std\left( a_{i} \right),
\end{equation}
where $N$, $a_{i}$, and $std$ represent the number of object classes, vector of varied affordance of an object (i.e., each column in Fig.~\ref{fig:fig3} (b)), and an operation to calculate the standard deviation of the non-zero values of a vector, respectively.

\begin{figure}[tb]
	\centering
    \includegraphics[width=0.8\linewidth]{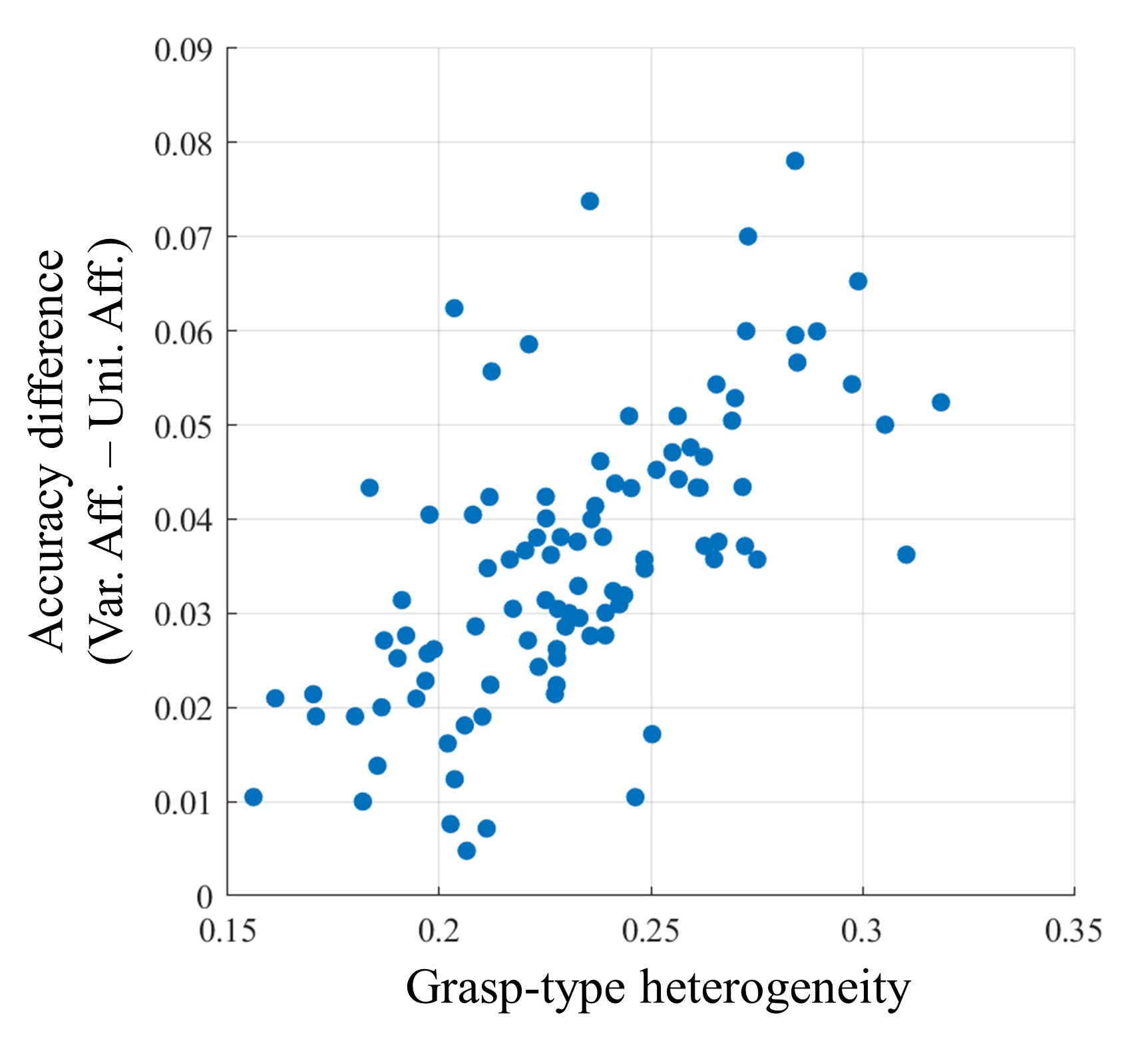}
	\caption{Performance difference between the pipelines plotted against grasp-type heterogeneity. Each plot represents a test dataset.}
	\label{fig:fig8}
\end{figure}

The sub-datasets were tested with the proposed pipeline using varied affordance and uniform affordance. The same CNN as in the comparison experiments was used.
Fig.~\ref{fig:fig8} shows the 
%performance difference of the pipelines
difference in performance between the two affordance types
plotted against the grasp-type heterogeneity. As hypothesized, the difference increased with the increasing grasp-type heterogeneity. This result indicates that the enhancing effect of the varied affordance is more pronounced when the degree of grasp-type heterogeneity is higher.

\subsection{Scenario 2: without real objects}\label{evaluation_pantomime}
In the previous section, we evaluated the pipeline for images of grasping real objects. On the contrary, robot teaching may not require real objects to be grasped in some situations (e.g., teaching in MR). In such situations, the captured images do not include real objects; however, a user can interact with an illusory object in MR (i.e., an MR object). Since such ``mimed'' images lack visual object information, image-based grasp-type recognition can become challenging. This section provides an evaluation of the performance of the proposed pipeline when mimed images and object affordance are available.

\subsubsection{Data preparation}\label{data_preparation_mime}
To obtain the CNN for recognizing the grasp types, we prepared a dataset of the mimed images captured by a HoloLens2 sensor \cite{hololens}. We used the texture-mapped 3D mesh models of the YCB objects described in Section \ref{data_preparation} as MR objects. Grasp achievement was determined by the type and number of the fingers in contact, following the definition in \cite{feix2015grasp}. The positions of the hand joints were estimated via the HoloLens2 API. During the collection of the images, a user grasped one of the rendered MR objects guided by visual cues that represent the contact state between the user's hand and MR object \cite{saito2021contact}. 
Among the object list in Fig.~\ref{fig:fig2}, 
the glass cleaner and the wine glass were ignored due to the lack of 3D models provided by \cite{calli2015ycb}, and the abrasive sponge was ignored due to the inability to express soft materials in MR.
Furthermore, ``small diameter'' grasping was ignored because of the difficulty in measuring the joint positions with the corresponding accuracy (i.e., within 1 cm \cite{feix2015grasp}). As the result of eliminating the ``small diameter'' grasping, we excluded objects with only one type of grasp (i.e., the pitcher and cooking skillet).

\begin{figure}[tb]
	\centering
	\includegraphics[scale=0.4]{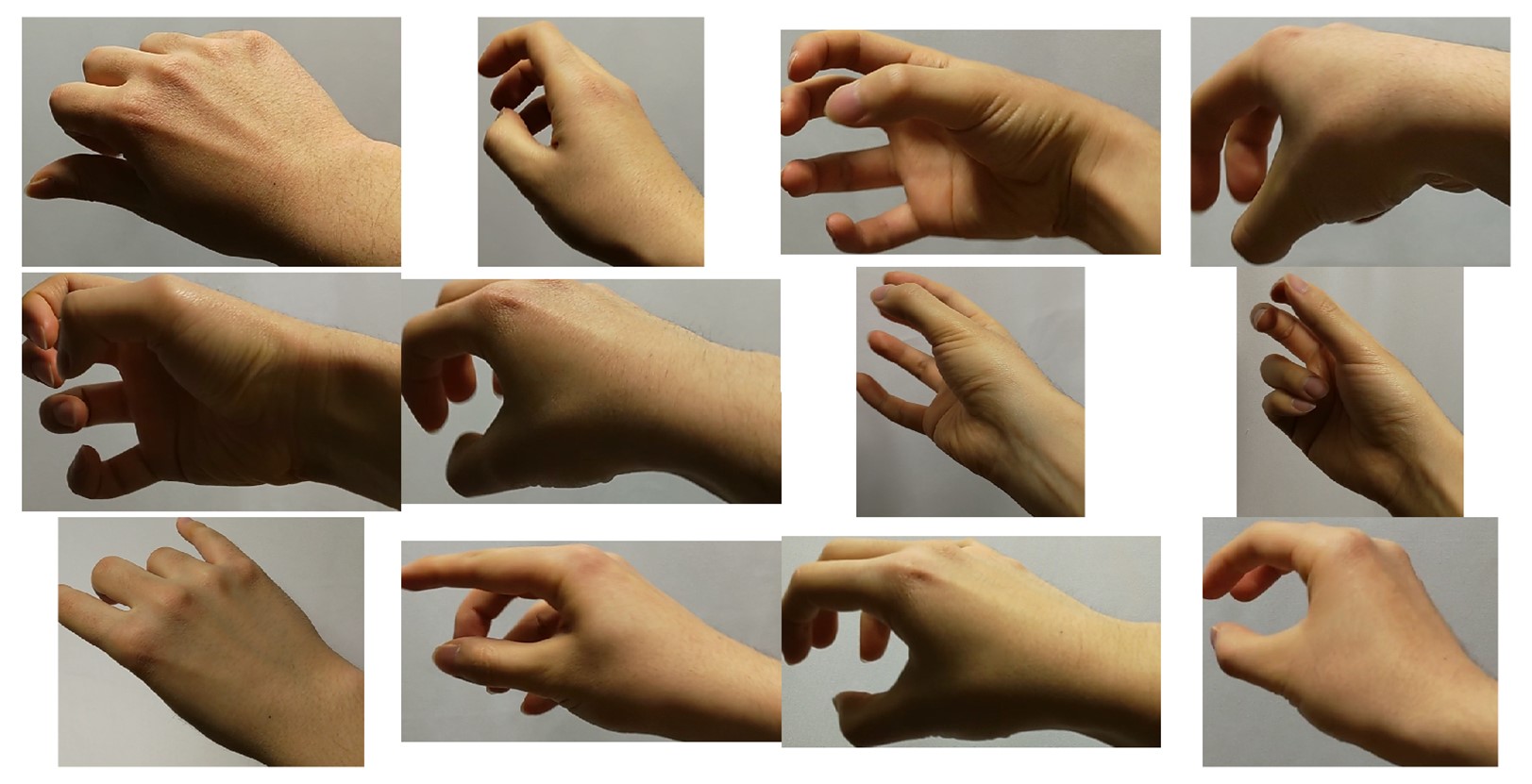}
	\caption{Examples of the mimed images captured by the HoloLens2 sensor. Although the grasped YCB objects were not captured, they were presented to the user in MR.}
	\label{fig:fig9B}
\end{figure}

Following the same recording and post-processing protocol described in Section \ref{data_preparation}, we collected a dataset containing 1000 mimed images for each object and grasp type (note that the MR objects were not captured in the images). We created two datasets under different lighting conditions and used one for training the CNN and the other for testing the pipeline. Fig.~\ref{fig:fig9B} shows examples of the images. 

\subsubsection{Effect of affordance on recognition}
\begin{figure}[tb]
	\centering
	\includegraphics[scale=0.32]{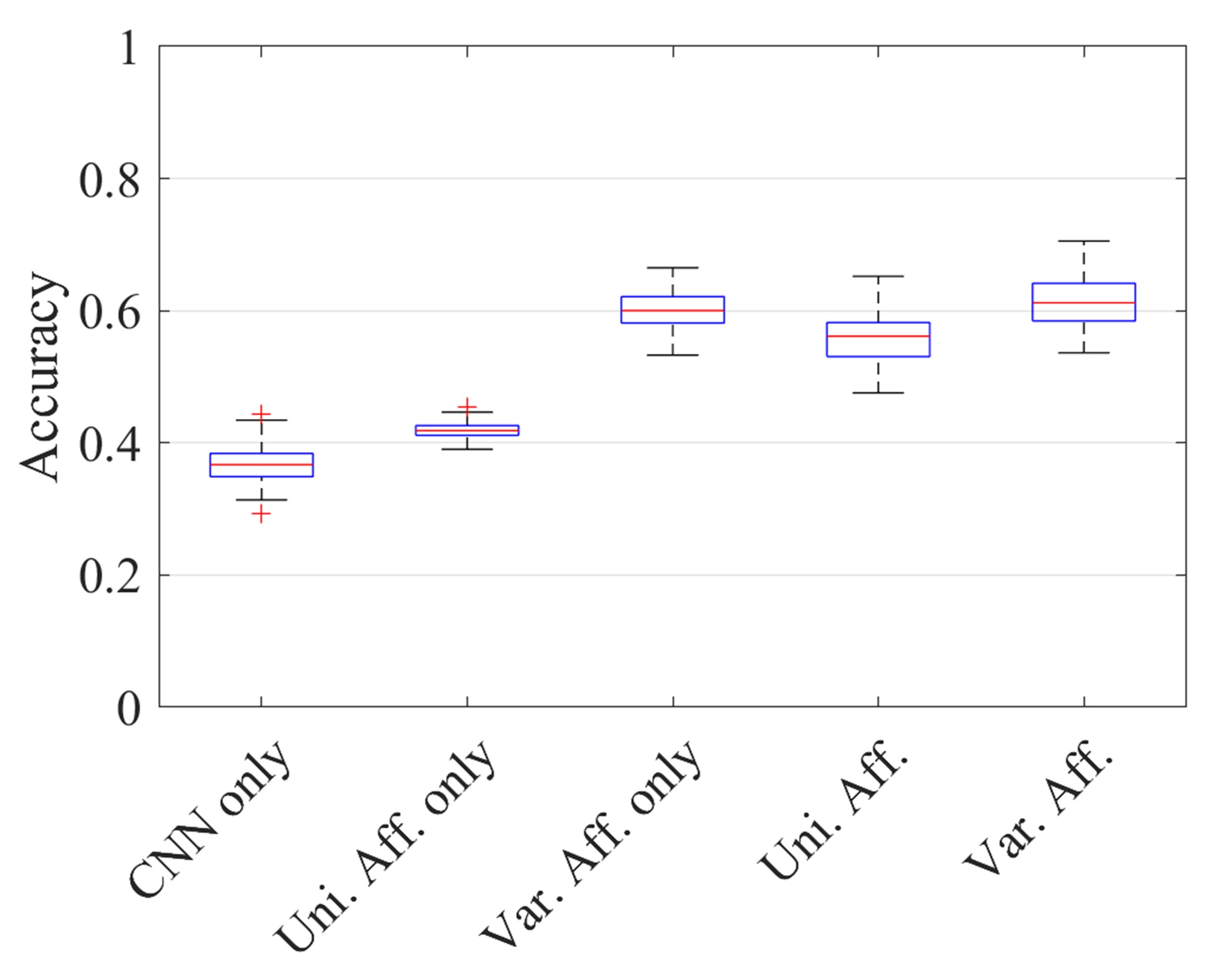}
	\caption{Performances of grasp-type recognition with different pipelines. The contractions are the same as in Fig. \ref{fig:fig5}.}
	\label{fig:fig10}
\end{figure}

We compared the same five methods as in Section \ref{scenario1}. The protocols to obtain the CNN and affordance database were the same as those described in Section \ref{cnn_affordance}. That is, we compared the performances of the five methods applied to a set of 100 test datasets.
Fig.~\ref{fig:fig10} shows the comparison results. Similar to the results in Section \ref{scenario1}, the proposed pipeline exhibited the highest performance. 
Although the CNN recognition for mimed images was inferior to recognition for real grasping images (see~Fig. \ref{fig:fig5}), the use of affordance proved to be effective. 

\begin{figure}[tb]
	\centering
    \includegraphics[width=\linewidth]{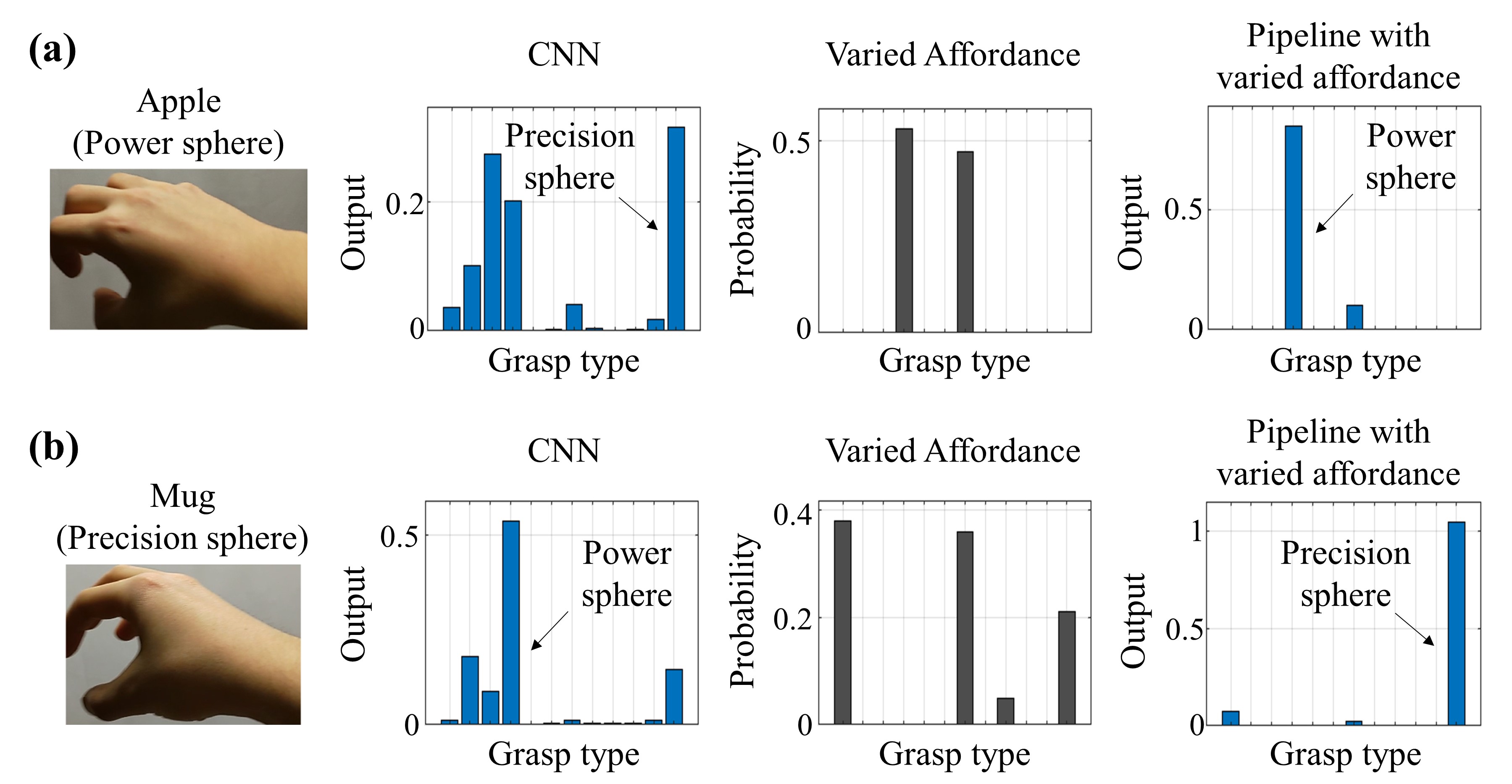}
	\caption{Examples of the CNN failed in recognizing the mimed images. (a) Recognition of ``power sphere'' grasping of an illusory apple. (b) Recognition of ``precision sphere'' grasping of an illusory mug. The order of grasp types is the same as in Fig. \ref{fig:fig2}, excluding ``small diameter'' grasping.}
	\label{fig:fig11}
\end{figure}

Additionally, we observed the two functions of object affordance (i.e., excluding the unlikely grasp types from the candidates and enhancing the likely grasp types among the candidates), similar to Section \ref{scenario1}. For example, Fig.~\ref{fig:fig11} shows cases where the CNN failed to discriminate between the ``power sphere'' and ``precision sphere,'' which appeared similarly in mimed grasping. Despite such similarity, the proposed pipeline using varied affordance succeeded by excluding either of them as candidates. 


\section{Discussion and conclusion} \label{conclusion}
\subsection{Summary of the experiments}
This study investigated the role of object affordance in guiding grasp-type recognition. To this end, we created two first-person image datasets containing images with and without the grasped objects, respectively. 
The results revealed the effects of object affordance in guiding CNN recognition: 1) it excludes the unlikely grasp types from the candidates and 2) enhances the likely grasp types among the candidates.
The enhancing effect was stronger when there was more heterogeneity between the likely grasp types. These findings suggest that object affordance can be effective in improving grasp-type recognition. 

The advantage of our proposed pipeline (Fig. \ref{fig:fig9A}) is that it can be updated independently of the CNN.
For example, if a user experiences a grasp type that is not assigned for an object, the pipeline can be updated by simply modifying the object affordance according to the user's feedback. 
As another example, if a user wants to interact with objects that are not registered in the affordance database, the pipeline can be updated by manually adding the object affordances. 
In the case of using uniform affordances, which showed promising results (Fig.~\ref{fig:fig5}), object affordances can be readily added by manually assigning the possible grasp types.
Such an approach is less expensive than updating a CNN by collecting a large number of grasping images depending on the use case. 

Recognition from the mimed images appears to be more difficult than that from the images of grasping real objects (Fig.~\ref{fig:fig5} and~\ref{fig:fig10}), indicating the importance of the presence of real objects in image-based recognition. The inferior performance with mimed images is reasonable because a grasp type depends on the shape of the hand and the fingers that are in contact with the object. Recognition from the mimed images may benefit from the findings of previous studies. For example, a study proposed combining other information, such as contact points and contact normals, which can be easily calculated for MR objects~\cite{aleotti2006grasp}. It may also be possible to utilize techniques developed in other research areas, such as sign language recognition for mimed images~\cite{al2021deep, wadhawan2021sign}. 
However, most importantly, object affordances can be applied to any recognition method that outputs a probability distribution (see equation~\ref{eq:affodance}). As long as visual ambiguities are inherently present (e.g., finger occlusion or absence of object), the proposed pipeline should be beneficial for grasp-type recognition.

\subsection{Methodological considerations}
The proposed pipeline used a text-based database that can be added and modified by the user according to each application (see Section~\ref{object_affordance}). This approach has two limitations. First, multiple object affordances cannot be associated with one text label. This becomes a problem when a user wants to register different object affordances for objects with the same name (e.g., grasp-type A for a \textit{cup} while grasp-type B for another \textit{cup}). Another limitation is that manual work is required to register object affordances. However, in practical use, we do not consider these characteristics to be a critical problem. Users can address the former issue by assigning different text labels to objects with different affordances. For the latter issue, we believe that the number of objects in the home environment is finite and falls within an acceptable range.

Regarding the system input, this study assumed that the pipeline can access the name of the grasped object and retrieve the affordance using the object name. For practical robot teaching applications, separate solutions to these requirements are required. To access the name of the grasped object, general object recognition or user input information can be used. For example, our robot teaching platform is designed to extract the name of the grasped object from human instructions \cite{wake2020verbal}. While this study used text matching to retrieve the affordance using object names, we could also employ a thesaurus or word embedding methods to cover the word variations.

The image datasets prepared in this study were collected against plain backgrounds under simple lighting conditions. Training images under a controlled environment often results in overfitting of a CNN and reduce its generalization performance. However, we consider the effect of using the controlled images to be limited for the following reasons. First, the CNNs were trained with cropped images with minimum background reflections (see Fig.~\ref{fig:fig2} and~\ref{fig:fig9B}). Second, to mitigate the lighting biases, we randomly shifted the colors of the training images. The effect of these pre-processing steps could be supported by the fact that the recognition performances of a single CNN were much higher than the chance rate (see Fig. \ref{fig:fig5} and~\ref{fig:fig10}). This study aimed to investigate the role of object affordance and did not scope the improvement of the generalization performance of CNNs; given the reasonable performance of the CNNs, the environmental condition was not critical to the paper's argument. 

\subsection{Future studies}
As future research, the proposed pipeline could be employed in a learning-from-observation (LfO) framework, where the object names can be estimated from verbal instructions. We are currently testing this hypothesis by integrating the pipeline with an LfO system that we developed in-house \cite{wake2020verbal,wake2021learning}.

\bibliographystyle{ieeetr}
\bibliography{mybibfile}

\end{document}